\documentclass[11pt,a4paper]{article}
\usepackage[hyperref]{emnlp2020}
\usepackage{times}
\usepackage{latexsym}

\aclfinalcopy
\usepackage{etoolbox}

\newtoggle{TEMPLATE}
\togglefalse{TEMPLATE}

\usepackage{microtype}

\usepackage{amsmath}

\usepackage{subfigure}
\usepackage{graphicx}
\usepackage{dirtytalk}

\title{Exploring Text Specific and Blackbox Fairness Algorithms \\in Multimodal Clinical NLP}

\font\authfont=cmr10.5 at 10pt 
\font\affilfont=cmr11 at 9pt

\author{
\authfont{
John Chen$^{1,2}$, 
Ian Berlot-Attwell $^{1,2}$, 
Safwan Hossain $^{1,2}$,
Xindi Wang $^{2,3}$,
Frank Rudzicz $^{1,2,4}$}
\\ 
\affilfont
$^1$ University of Toronto, 
$^2$ Vector Institute,
$^3$ University of Western Ontario, 
$^4$ St. Michael's Hospital
\\
  \small{\texttt{\{johnc, ianberlot, hossa120, frank \}@cs.toronto.edu}}\\ 
 \small{\texttt{xwang842@uwo.ca}} 
}

\date{}

\usepackage{todonotes} 
\usepackage{comment}
\newif\ifshow 
\showfalse 

\ifshow
  \includecomment{wrap}
\else
  \excludecomment{wrap} 
\fi

\begin{document}
\maketitle
\begin{abstract}

Clinical machine learning is increasingly multimodal, collected in both structured tabular formats and unstructured forms such as free text.  We propose a novel task of exploring \textit{fairness} on a multimodal clinical dataset, adopting \textit{equalized odds} for the downstream medical prediction tasks. To this end, we investigate a modality-agnostic fairness algorithm - equalized odds post processing - and compare it to a text-specific fairness algorithm: debiased clinical word embeddings. Despite the fact that debiased word embeddings do not explicitly address equalized odds of protected groups, we show that a  text-specific approach to fairness may simultaneously achieve a good balance of performance \textit{and} classical notions of fairness. We hope that our paper inspires future contributions at the critical intersection of clinical NLP and fairness. The full source code is available here: \url{https://github.com/johntiger1/multimodal_fairness}

\end{abstract}


\section{Introduction}

Natural language processing is increasingly leveraged in sensitive domains like healthcare. For such critical tasks, the need to prevent discrimination and bias is imperative. Indeed, ensuring equality of health outcomes across different groups has long been a guiding principle of modern health care systems \cite{culyer1993equity}.
Moreover, medical data presents a unique opportunity to work with different \textit{modalities}, specifically \textit{text} (e.g., patient narratives, admission notes, and discharge summaries) and numerical or categorical data (often denoted \textit{tabular} data, e.g., clinical measurements such as blood pressure, weight, or demographic information like ethnicity). Multi-modal data is not only reflective of many real-world settings, but machine learning models which leverage both structured and unstructured data often achieve greater performance than their individual constituents \cite{horng2017creating}. While prior work studied fairness in the text and tabular modalities in isolation, there is little work on applying notions of algorithmic fairness in the broader multimodal setting \citep{10.1145/3368555.3384448, chen2018my}. 






Our work brings a novel perspective towards studying fairness algorithms for models which operate on \textit{both} text and tabular data, in this case applied to the MIMIC-III clinical dataset (MIMIC-III) \cite{MIMIC}. We evaluate two fairness algorithms: equalized-odds through post-processing, which is agnostic to the underlying classifier, and word embedding debiasing which is a text-specific technique. We show that ensembling classifiers trained on structured and unstructured data, along with the aforementioned fairness algorithms, can both improve performance and mitigate unfairness relative to their constituent components. We also achieve strong results on several MIMIC-III clinical benchmark prediction tasks using a dual modality ensemble; these results may be of broader interest in clinical machine learning \citep{Harutyunyan2019, khadanga2019using}. 




\section{Background}



\subsection{Combining Text and Tabular Data in Clinical Machine Learning}
Prior work has shown that combining unstructured text with vital sign time series data improves performance on clinical prediction tasks. \citet{horng2017creating} showed that augmenting an SVM with text information in addition to vital signs data improved retrospective sepsis detection. \citet{akbilgic2019unstructured} showed that using a text-based risk score improves performance on prediction of death after surgery for a pediatric dataset. Closest to our work, \citet{khadanga2019using}  introduced a joint-modality neural network which outperforms single-modality neural networks on several benchmark prediction tasks for MIMIC-III. 

\subsection{Classical fairness metrics}
Many algorithmic fairness notions fall into one of two broad categories: individual fairness enforcing fairness across individual samples, and group fairness seeking fairness across protected groups (e.g. race or gender). 
We focus on a popular group-level fairness metric: {\em Equalized Odds} (EO) \citep{NIPS2016_6374}. Instead of arguing that average classification probability should be equal across all groups (also known as {\em Demographic Parity}) -- which may be unfair if the underlying group-specific base rates are unequal -- EO allows for classification probabilities to differ across groups only through the underlying ground truth. Formally, a binary classifier $\widehat{Y}$ satisfies EO for a set of groups $\mathcal{S}$ if, for ground truth $Y$ and group membership $A$:
\begin{equation*}\label{EOeq}\small
    \begin{split}
        \text{Pr} ( \hat{Y}=1 \,|\, Y = y, A=a ) = \text{Pr} ( \hat{Y}=1 \,|\, Y = y, A=a' )\\ \forall y \in \{0,1\}, \forall a,a' \in \mathcal{S}
    \end{split}
\end{equation*}

In short, the true positive (TP) and true negative (TN) rates should be equal across groups. 

\subsection{Equalized Odds Post Processing}
\citet{NIPS2016_6374} proposed a model-agnostic post-processing algorithm that minimizes this group specific error discrepancy while considering performance. Briefly, the post-processing algorithm determines group-specific random thresholds based on the intersection of group-specific ROC curves. The multi-modality of our underlying data and the importance of privacy concerns in the clinical setting make post-processing especially attractive as it allows fairness to be achieved agnostic to the inner workings of the base classifier. 


\subsection{Debiasing word embeddings}

Pretrained word embeddings encode the societal biases of the underlying text on which they are trained, including gender roles  and racial stereotypes
\citep{bolukbasi2016man, zhao-etal-2018-learning,manzini-etal-2019-black}. 
Recent work has attempted to mitigate this bias in context-free embeddings while preserving the utility of the embeddings.
\citet{bolukbasi2016man} analyzed gender subspaces by comparing distances between word vectors with pairs of gender-specific words to remove bias from gender-neutral words. \citet{manzini-etal-2019-black} extended this work to the multi-class setting, enabling debiasing in race and religion. Concurrent to their work, \cite{ravfogel2020null} propose  iterative null space projection as a technique to hide information about protected attributes by casting it into the null space of the classifier.  Following the recent popularity of BERT and ELMo, \citet{liang2020towards} consider extending debiasing to sentence-level, contextualized representations. 




\begin{figure}
    \centering
    \includegraphics[width=\linewidth]{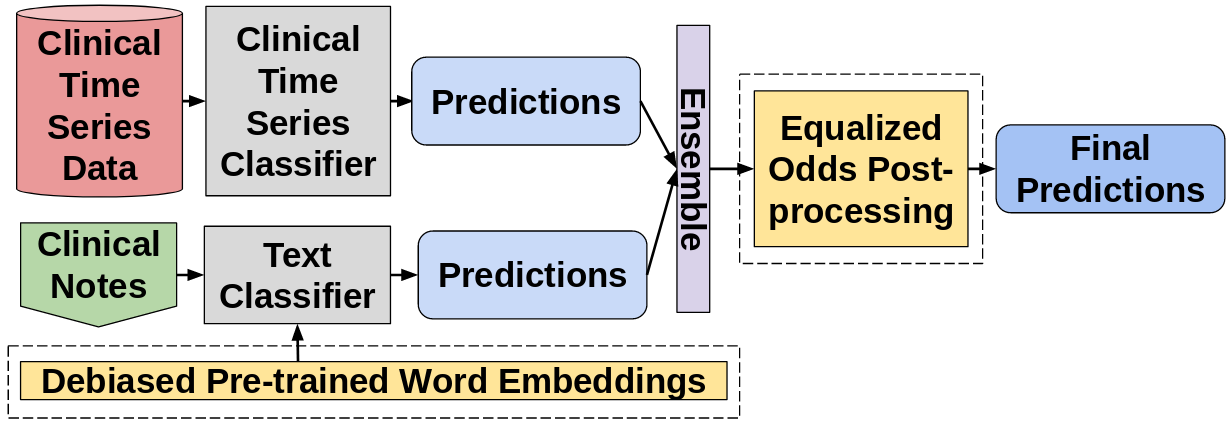}
    \caption{Experimental setup and ensemble architecture. Fairness approaches are indicated in dotted boxes. 
    }
    \label{fig:architecture}
\end{figure}

\section{Experimental Setup}\label{sec:experimental_setup}

\subsection{Clinical Prediction Tasks} \label{clinical_pred_tasks}
\begin{figure}[h!]
    \centering
    \includegraphics[width=\linewidth]{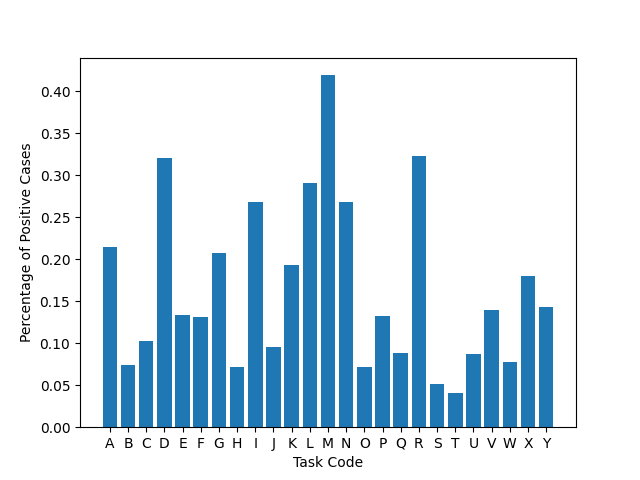}
    \caption{Percentage of positive train cases for each of the 25 phenotyping tasks. The critical care conditions corresponding to the task codes can be found in Table \ref{tab:phen_perc_legend} of the Appendix}
    \label{fig:phenotype_perc_tr}
\end{figure}

MIMIC-III contains deidentified health data associated with ~60,000 intensive care unit (ICU) admissions \citep{MIMIC}. It contains  both unstructured textual data (in the form of clinical notes) and structured data (in the form of clinical time series data  and demographic, insurance, and other related meta-data). 
We focus on two benchmark binary prediction tasks for ICU stays previously proposed by \citet{Harutyunyan2019}: in-hospital mortality prediction (IHM), which aims to predict mortality based on the first 48 hours of a patient's ICU stay, and phenotyping, which aims to retrospectively predict the acute-care conditions that impacted the patient.  Following \citet{khadanga2019using} we extend the prediction tasks to leverage clinical text linked to their ICU stay. For both tasks the classes are higly imbalanced: in the IHM task only 13.1\% of training examples are positive, and the relative imbalance of the labels in the phenotyping class can be seen in Figure \ref{fig:phenotype_perc_tr}. To account for the label imbalance we evaluate performance using AUC ROC and AUC PRC.  More details can be found in Appendix \ref{appendix:a}. 

\subsection{Fairness Definition}
Next, we consider how we can extend a definition of fairness to this multimodal task. Following work by \citet{10.1145/3368555.3384448} in the single-modality setting, we examine True Positive and True Negative rates on our clinical prediction task between different protected groups. Attempting to equalize these rates corresponds to satisfying \emph{Equalized Odds}. EO satisfies many desiderata within clinical settings, and has been used in previous clinical fairness work \citep{10.1145/3306618.3314278, doi:10.1111/j.1468-2850.1997.tb00104.x, pmlr-v106-pfohl19a}. While EO does not explicitly incorporate the \textit{multimodality} of our data, it accurately emphasizes the importance of the \textit{downstream} clinical prediction task on the protected groups.   Nonetheless, we acknowledge that EO alone is insufficient for practical deployment; na\"ive application can result in unacceptable performance losses and thus consultations with physicians and stakeholders must be held \citep{Rajkomar2018}.

\subsection{Classification Models}

We provide brief descriptions below with details available in Appendix \ref{appendix:b}.
\begin{itemize}
\setlength{\itemsep}{-0.9pt}
\item \textbf{Structured Data Model: }Following \citet{Harutyunyan2019}, we use a channel-wise bidirectional Long Short Term Memory network (bi-LSTM).  
\item \textbf{Unstructured Textual Data:} We use a CNN encoder to extract the semantic features from clinical notes. Importantly, we experiment with training word embeddings from scratch and utilizing pre-trained BioWordVec embeddings \citep{zhang2019biowordvec}. 


\item \textbf{Ensemble:} We perform logistic regression on the output binary classification probabilities from the previous models. 

\end{itemize}
\section{Fairness Setup}




\subsection{Sensitive groups}
Recall that EO explicitly ensures fairness with respect to sensitive groups while debiasing implicitly depends upon it. Leveraging the demographic data in MIMIC-III, we consider ethnicity (divided into Asian, Black, Hispanic, White and other), biological sex (divided into male and female), and insurance type (divided into government, medicare, medicaid, self-pay, private, and unknown). With the exception of biological sex, the sensitive groups are highly imbalanced (see Table \ref{tab:in_hosp_mort_groups}). Note that insurance-type has been shown to be a proxy for socioeconomic status (SES) \cite{pmid30794127}. 

\begin{table}[]
    \centering
    
    \begin{tabular}{|c|c|c|c|}
    \hline
    {\begin{tabular}[c]{@{}c@{}}Sensitive \\ Group \end{tabular}}     &  {\begin{tabular}[c]{@{}c@{}}Train \\ Count\end{tabular}} & {\begin{tabular}[c]{@{}c@{}} Test \\ Count \end{tabular}} & \% of Test \\
    \hline
    F & 7940 & 1415 & 44.0 \% \\
    M & 9708 & 1778 & 56.0 \% \\
    \hline
    ASIAN & 408 & 60 & 1.9 \% \\
    BLACK & 1658 & 285 & 8.9 \% \\
    HISPANIC & 521 & 107 & 3.3 \% \\
    OTHER & 2655 & 459 & 14.4 \% \\
    WHITE & 12406 & 2282 & 71.5 \% \\
    \hline
    Government & 356 & 74 & 2.3 \% \\
    Medicaid & 1362 & 205 & 6.4 \% \\
    Medicare & 9857 & 1757 & 55.0 \% \\
    Private & 4946 & 932 & 29.2 \% \\
    Self Pay & 133 & 33 & 1.0 \% \\
    UNKNOWN & 994 & 192 & 6.1 \% \\
    \hline
    \end{tabular}
    
    \caption{Distribution of sensitive-attributes over train and test data for the In-Hospital Mortality task}
    \label{tab:in_hosp_mort_groups}
\end{table}

\subsection{Equalized Odds Post-Processing}
We apply our equalized-odds post processing algorithm on the predictions of the trained single-modality  classifiers (physiological signal LSTM model as well as text-only CNN model) as well as the trained ensemble classifier. Note that we apply EO postprocessing only once for each experiment: either on the outputs of the single-modality model, or on the ensemble predictions. 
 The fairness approaches are mutually exclusive: we do not consider applying EO postprocessing together with debiased word embeddings. We consider using both soft prediction scores (interpretable as probabilities) as well as thresholded hard predictions as input to the post-processing algorithm. These choices impact the fairness performance trade-off as discussed further in Section \ref{sec:Results}.


\subsection{Socially Debiased Clinical Word Embeddings}

While clinically pre-trained  word embeddings may improve downstream task performance, they are not immune from societal bias \cite{khattak2019survey}.  
We socially debias these clinical word embeddings following \citet{manzini-etal-2019-black}. We manually select sets of social-specific words (see Appendix \ref{appendix:c}) to identify the fairness-relevant social bias subspace. 
Formally, having identified the basis vectors $\{b_{1}, b_{2}, ..., b_{n}\}$ of the social bias subspace $\mathcal{B}$, we can find the projection $w_B$ of a word embedding $w$:
\[w_\mathcal{B} = \sum_{i=1}^{n} \langle{w}, b_{i}\rangle{b_{i}}\]

Next we apply hard debiasing, which will remove bias from existing word embeddings by subtracting $w_B$, their component in this fairness subspace. This yields $w'$, our socially debiased word embedding: 

\begin{align*}
    w' = \frac{w - w_B}{\|w-w_B\|}
\end{align*}





We consider debiasing with respect to race and gender. The race debiased embeddings are re-used for insurance tasks as empiric research has indicated that the use of proxy groups in fairness can be effective \citep{DBLP:journals/corr/abs-1806-11212} and SES is strongly related to race \citep{Williams2016}. 



\section{Results and Analysis}\label{sec:Results}

\begin{table}[h]
\centering
\resizebox{\columnwidth}{!}{
\begin{tabular}{clllll}
\cline{2-5}
\multicolumn{1}{l|}{} &
  \multicolumn{2}{c|}{IHM} &
  \multicolumn{2}{c|}{Phenotyping} &
  \multicolumn{1}{l}{} \\ \cline{1-5}
\multicolumn{1}{|l|}{} &
  \multicolumn{1}{c|}{\begin{tabular}[c]{@{}c@{}}AUC \\ PRC\end{tabular}} &
  \multicolumn{1}{c|}{\begin{tabular}[c]{@{}c@{}}AUC \\ ROC\end{tabular}} &
  \multicolumn{1}{c|}{\begin{tabular}[c]{@{}c@{}}Macro \\ AUCROC\end{tabular}} &
  \multicolumn{1}{c|}{\begin{tabular}[c]{@{}c@{}}Overall \\ AUCROC\end{tabular}} &
   \\ \cline{1-5}
   \multicolumn{1}{|c|}{\begin{tabular}[c]{@{}c@{}}Harutyunyan et. al\\  (2019) -- No Text\end{tabular}}& \multicolumn{1}{l|}{0.515} & \multicolumn{1}{l|}{0.862} &  
\multicolumn{1}{c|}{0.776} &
\multicolumn{1}{c|}{0.825} &  \\ \cline{1-5}
\multicolumn{1}{|c|}{\begin{tabular}[c]{@{}c@{}}Khadanga et. al \\  (2019) -- Ensemble\end{tabular}}    & \multicolumn{1}{l|}{0.525} & \multicolumn{1}{l|}{0.865} &
\multicolumn{1}{c|}{--} & 
\multicolumn{1}{c|}{--} &   \\ \cline{1-5}
\multicolumn{1}{|c|}{\begin{tabular}[c]{@{}c@{}}Ours -- Text Only\end{tabular}} &
  \multicolumn{1}{l|}{0.472} &
  \multicolumn{1}{l|}{0.815} &
\multicolumn{1}{c|}{0.766} &  
\multicolumn{1}{c|}{0.829} & \\ \cline{1-5}
   \multicolumn{1}{|c|}{\begin{tabular}[c]{@{}c@{}}Ours -- Text Only \\ + BioWordVec \end{tabular}} &
  \multicolumn{1}{l|}{0.489} &
  \multicolumn{1}{l|}{0.841} &
\multicolumn{1}{c|}{0.771} & 
\multicolumn{1}{c|}{0.837} &  \\ \cline{1-5}
\multicolumn{1}{|c|}{Ours -- Ensemble}           & \multicolumn{1}{l|}{\textbf{0.582}} & 
\multicolumn{1}{l|}{0.880} &  
\multicolumn{1}{c|}{0.822} &
\multicolumn{1}{c|}{0.861} &  \\ \cline{1-5}
\multicolumn{1}{|c|}{\begin{tabular}[c]{@{}c@{}}Ours -- Ensemble \\ + BioWordVec\end{tabular}}           & \multicolumn{1}{l|}{\textbf{0.582}} & \multicolumn{1}{c|}{\textbf{0.886}} &  
\multicolumn{1}{c|}{\textbf{0.829}} & 
\multicolumn{1}{c|}{\textbf{0.870}} & \\ \cline{1-5}
\end{tabular}}
\caption{Leveraging clinical pretrained word embeddings improves performance compared to training word embeddings from scratch in the text-only model. Ensembling the text-only model with the clinical time series classifier improves performance further. 
 }
\label{tab:my-table}
\end{table}



\begin{figure*}[htp]
  \centering
  \includegraphics[width=\linewidth]{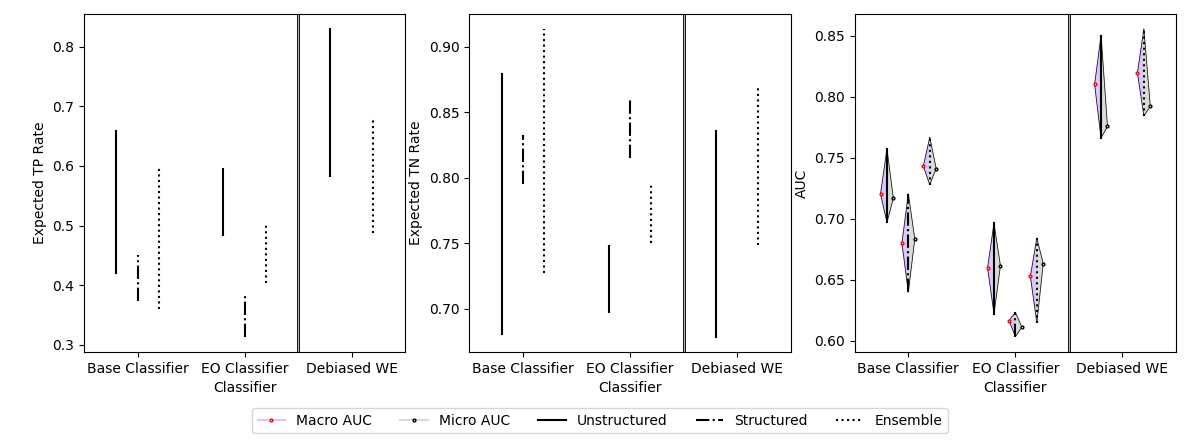}
  \caption{  
  Plots of TP Rate, TN Rate, and AUC on phenotyping task M for groups defined by sensitive attribute of race. Each vertical black line represents a classifier (line style indicating modality); the length of the line represents the range of scores over fairness groups. In the TP/TN graphs, a shorter line represents better fairness; there is less discrepancy between the maximum and minimum group-specific TP/TN rates. In the AUC graph (far right), the higher the vertical position of the line, the better the performance. EO is effective at reducing the spread in TP/TN rates for the ensemble classifier (first two graphs) at the cost of performance (far right) graph. Meanwhile, debiased word embeddings both improves fairness, reducing the length of the line in the first two graphs, while achieving superior performance in AUC graph  }\label{fig:big_plot}
  
  

  %
\end{figure*}

\subsection{Ensembling clinical word embeddings with structured data improves performance} 
 
Empirically, we observe superior performance to prior literature on a suite of clinical prediction tasks in Table \ref{tab:my-table}; more tasks are evaluated in Appendix Table \ref{appendix:a}. Full hyperparameter settings and code for reproducibility can be found here \footnote{\url{https://github.com/johntiger1/multimodal_fairness/clinicalnlp}}. The ensemble model 
outperforms both constituent classifiers (AUC plot on Figure \ref{fig:big_plot}). This holds even when fairness/debiasing techniques are applied, emphasizing the overall effectiveness of leveraging multi-modal data. However, the ensemble's improvements in performance do not directly translate to improvements in fairness; see the True Positive (TP) graph in Figure \ref{fig:big_plot}, where the maximum TP gap remains consistent under the ensemble.



\subsection{Debiased word embeddings and the fairness performance trade-off}
Improving fairness usually comes at the cost of reduced performance \citep{pmlr-v81-menon18a}. Indeed, across all tasks, fairness groups and classifiers, we observe the group-specific disparities of TP and TN rates generally diminish when equalized odds post-processing is used (see Appendix \ref{sec:appendix:full_result} for additional results). However, this post-processing also leads to a degradation in the AUC. 
Note that we apply EO-post processing on hard (thresholded) predictions of the classifiers. If instead \textit{soft} prediction scores are used as inputs to the post-processing step, both the performance degradation and the fairness improvement are softened \citep{NIPS2016_6374}. 




Generally, word embedding debiasing (WED) also helps reduce TP/TN discrepancies, although not to the same extent as EO postprocessing. Remarkably, in certain tasks, WED also yields a performance improvement, even compared to the fairness-free, unconstrained ensemble classifier. In particular, for the AUC graph in Figure \ref{fig:big_plot}, leveraging debiased word embeddings improves the performance of the ensemble; at the same time, the TP and TN group discrepancy ranges are improved.  However, we stress that this outcome was not consistently observed and further investigation is warranted.



We emphasize that EO and WED serve different purposes with different motivations. While EO explicitly seeks to minimize the TP/TN range between sensitive groups (reflected in its performance on the first two plots in Figure \ref{fig:big_plot}), WED seeks to neutralize text-specific bias in the word-embeddings. Despite the difference in goals, and despite operating only on the text-modality of the dataset, WED is still able to reduce the group-specific TP/TN range; recent work on \textit{proxy fairness} in text has shown that indirect correlation between bias in text and protected attributes may be useful in achieving parity \citep{romanov2019s}.


Although WED demonstrate some good properties with respect to both fairness and performance for our specific dataset and task, we caution that they represent only one approach to fairness in NLP \cite{blodgett2020language}. Indeed, WED suffers from shortcomings related to intersectional fairness \citep{gonen2019lipstick}, and we encourage further discussion into concretely defining fair, real-world NLP tasks and developing novel algorithms. 

Our results highlight the important role practitioners and stakeholders play in algorithmic fairness on clinical applications. The trade-off between performance and fairness, whether between the soft and hard labels used for EO, or between EO and debiased word embeddings, must be balanced based on numerous real world factors. 

\section{Discussion}

 
 In this paper, we propose a novel multimodal fairness task for the MIMIC-III dataset, based on equalized odds. We provide two baselines: a classifier-agnostic fairness algorithm (equalized odds post-processing) and a text-specific fairness algorithm (debiased word embeddings). We observe that both methods generally follow the fairness performance tradeoff seen in single-modality tasks. EO is more effective at reducing the disparities in group-specific error rates while word-embedding debiasing has better performance. 
 Future work can consider more generalized notions of fairness such as preferences-based frameworks, or extend text-specific fairness to contextualized word embeddings \citep{hossain2020designing, 10.1145/3368555.3384448}. 
 Further analysis of the fairness performance tradeoff, especially in multimodal settings, will facilitate equitable decision making in the clinical domain.

 \section{Acknowledgements}
 We would like to acknowledge Vector Institute for office and compute resources. We would also like to thank Matt Gardner for his help with answering questions when using AllenNLP \cite{Gardner2017AllenNLP}. John Chen and Safwan Hossain are funded by an Ontario Graduate Scholarship and a Vector Institute Research Grant. Ian Berlot-Attwell is funded by a Canada Graduate Scholarships-Master’s, and a Vector Institute Research Grant. Frank Rudzicz is supported by a CIFAR Chair in AI.

\bibliography{anthology,emnlp2020}
\bibliographystyle{acl_natbib}

\clearpage
\appendix

\section{Details on Clinical Prediction Tasks}






\label{appendix:a}


\subsection{Defining the Multimodal MIMIC-III Benchmark Prediction Tasks}
Existing work by \cite{Harutyunyan2019} previously defined four benchmark clinical prediction tasks on ICU stays information from the large MIMIC-III database. They produce a derived dataset, focusing on 17 timeseries clinical features, without text. The goal is predict the task specific outcome (mortality, phenotyping, decompensation, length-of-stay) for the given ICU stay.  We utilize their derived dataset directly, which provides training and test examples for all four tasks, but join the derived dataset back with the original to obtain linked clinical text. We make the key choice that we drop examples without \textit{relevant} (i.e. no causal leakage) extracted clinical notes, as in \cite{khadanga2019using}. Thus, we concretely define the Combined Modality MIMIC-III Benchmark Prediction Task as extending the benchmark clinical prediction task by \cite{Harutyunyan2019} to include linked clinical text. If there are no notes associated with an example, then we remove this instance from the task. Note that we also drop ICU stays which only have unusable notes due to causal leakage; for instance death reports for mortality prediction. 





\subsection{Note extraction}
To extract relevant notes, we build a mapping from the derived dataset provided by \cite{Harutyunyan2019} and the MIMIC-III database. For each training and test instance in each task, we find the clinical notes in the MIMIC-III database. For the IHM task, if we do not find any notes within the first 48 hours of their stay, we drop the patient, since there is no \textit{relevant} textual information. Note that this is consistent with the original task formulation by \cite{Harutyunyan2019} of in-hospital mortality prediction using at most the first 48 hours of clinical data. Furthermore, this follows \cite{khadanga2019using}. 

For the phenotyping task, which is not covered by \cite{khadanga2019using}, we relax this time condition. In the original formulation of the task, phenotyping is a \textit{retrospective} multilabel multiclass classification task, meaning that all vital signs data associated with the ICU stay is provided and can be used by the model. Therefore, we only drop the patient if there are no notes for the entire ICU stay.


\subsection{Preprocessing}
We use the same preprocessing as in \cite{khadanga2019using}, finding it to be mildly beneficial for performance. 

\subsection{Cohort statistics}
In the medical literature, \textit{cohort selection} is the process of selecting the population of patients for inclusion in a study. These patients will then provide the training instances for the clinical prediction task.  We report the cohort statistics for our binary clinical prediction multimodal tasks.

\label{sec:appendix:cohort_stats}
\subsubsection{In-Hospital Mortality}

\begin{tabular}{|c|c|c|c|}
\hline
{\begin{tabular}[c]{@{}c@{}}Sensitive \\ Group \end{tabular}}     &  {\begin{tabular}[c]{@{}c@{}}Train \\ Count\end{tabular}} & {\begin{tabular}[c]{@{}c@{}} Test \\ Count \end{tabular}} & \% of Test \\
\hline
F & 7940 & 1415 & 44.0 \% \\
M & 9708 & 1778 & 56.0 \% \\
\hline
ASIAN & 408 & 60 & 1.9 \% \\
BLACK & 1658 & 285 & 8.9 \% \\
HISPANIC & 521 & 107 & 3.3 \% \\
OTHER & 2655 & 459 & 14.4 \% \\
WHITE & 12406 & 2282 & 71.5 \% \\
\hline
Government & 356 & 74 & 2.3 \% \\
Medicaid & 1362 & 205 & 6.4 \% \\
Medicare & 9857 & 1757 & 55.0 \% \\
Private & 4946 & 932 & 29.2 \% \\
Self Pay & 133 & 33 & 1.0 \% \\
UNKNOWN & 994 & 192 & 6.1 \% \\
\hline
\end{tabular}

\subsubsection{Phenotyping}

\begin{tabular}{|c|c|c|c|}
\hline
{\begin{tabular}[c]{@{}c@{}}Sensitive \\ Group \end{tabular}}     &  {\begin{tabular}[c]{@{}c@{}}Train \\ Count\end{tabular}} & {\begin{tabular}[c]{@{}c@{}} Test \\ Count \end{tabular}} & \% of Test \\
\hline
F & 15638 & 2750 & 44\% \\
M & 19803 & 3504 & 56\% \\
\hline
ASIAN & 826 & 133 & 2.1 \% \\
BLACK & 3378 & 575 & 9.1 \% \\
HISPANIC & 1158 & 206 & 3.3 \% \\
OTHER & 5004 & 854 & 13.7 \% \\
WHITE & 25075 & 4486 & 71.7 \% \\
\hline
Government & 845 & 150 & 2.4 \% \\
Medicaid & 2850 & 433 & 6.9 \% \\
Medicare & 18702 & 3298 & 52.7 \% \\
Private & 10784 & 1923 & 30.7 \% \\
Self Pay & 380 & 73 & 1.2 \% \\
UNKNOWN & 1880 & 377 & 6.0 \% \\
\hline
\end{tabular}


\subsection{Task Statistics}

\subsubsection{In-Hospital Mortality}

\begin{tabular}{|c|c|c|}
\hline
Label & Train Set Count & Test Set Count  \\
\hline
 0 & 15337   & 2829 \\
 1 & 2311   & 364 \\
\hline
\end{tabular}

\subsubsection{Phenotyping}

\begin{figure}[h!]
    \centering
    \includegraphics[width=\linewidth]{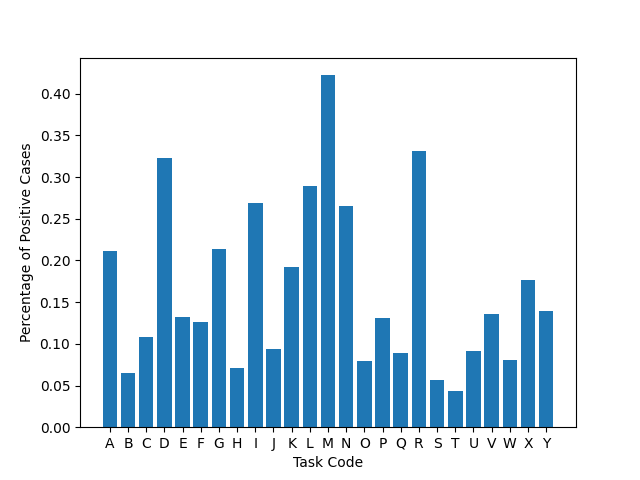}
    \caption{Percentage of positive test cases for each of the 25 phenotyping tasks}
    \label{fig:phenotype_perc_te}
\end{figure}

Plots of the prevalance of the 25 critical care conditions can be found in Figures \ref{fig:phenotype_perc_te} and \ref{fig:phenotype_perc_tr} for the test and train sets respectively, a legend that doubles as the full list of phenotyping tasks is available in Table \ref{tab:phen_perc_legend}.

\begin{table}[]
    \centering
    \begin{tabular}{p{1cm}p{5cm}}
    \hline
     Code   & Task                                                                           \\
    \hline
     A      & Acute and unspecified renal failure                                            \\
     B      & Acute cerebrovascular disease                                                  \\
     C      & Acute myocardial infarction                                                    \\
     D      & Cardiac dysrhythmias                                                           \\
     E      & Chronic kidney disease                                                         \\
     F      & Chronic obstructive pulmonary disease and bronchiectasis                       \\
     G      & Complications of surgical procedures or medical care                           \\
     H      & Conduction disorders                                                           \\
     I      & Congestive heart failure                                                       \\
     J      & Coronary atherosclerosis and other heart disease                               \\
     K      & Diabetes mellitus with complications                                           \\
     L      & Diabetes mellitus without complication                                         \\
     M      & Disorders of lipid metabolism                                                  \\
     N      & Essential hypertension                                                         \\
     O      & Fluid and electrolyte disorders                                                \\
     P      & Gastrointestinal hemorrhage                                                    \\
     Q      & Hypertension with complications and secondary hypertension                     \\
     R      & nonhypertensive                                                                \\
     S      & Other liver diseases                                                           \\
     T      & Other lower respiratory disease                                                \\
     U      & Other upper respiratory disease                                                \\
     V      & Pleurisy                                                                       \\
     W      & Pneumonia (except that caused by tuberculosis or sexually transmitted disease) \\
     X      & pneumothorax                                                                   \\
     Y      & pulmonary collapse                                                             \\
    \hline
    \end{tabular}
    \caption{List of critical care conditions in the phenotyping task, and their corresponding alphabetic codes.}
    \label{tab:phen_perc_legend}
\end{table}

\section{Model Details}
\label{appendix:b}
\subsection{Structured Data Model}
We use the baseline developed by \citet{Harutyunyan2019}. The structured data model  takes as input a time-series of 17 clinical variables, which are extracted features for the benchmark tasks introduced in the same paper. The model is a channel-wise LSTM where each clinical variable is transcoded by a bidirectional LSTM, concatenated with the other transcoded sequences and passed to a final LSTM for prediction.
\subsection{Unstructured Data Model}
We implement a simple CNN-based encoder \cite{kim2014convolutional, zhang2015sensitivity} to process the clinical notes and produce a task-specific prediction. We experiment with various settings including model architecture, word embedding dimension, preprocessing, varying the maximum number of tokens, L2 regularization and batch size. Below, we report the final hyperparameters and settings used to generate all plots and reported throughout. 

Our CNNEncoder is built using the AllenNLP framework \citep{Gardner2017AllenNLP}. We use 1D kernel (\textit{n}-gram) filter sizes of 2, 3 and 5, learning 5 filters for each filter size. Convolution is done on word embedding representations of the input, across \textit{n}-gram windows of the sequence, and are pooled before being combined. The CNNEncoder produces a single fixed size vector, and we use a simple linear layer on top to perform the classification.  

For all multimodal tasks, we limit the maximum number of tokens input to 1536, taking the most recent notes first (taking care to avoid causal leakage as described in \ref{clinical_pred_tasks}), and apply preprocessing as in \cite{khadanga2019using}. For the decompensation task, we subsample the number of training instances due to engineering and efficiency reasons. From 2 million possible training instances, we sample 50 000 examples, with weighting to balance the number of positive and negatively training instances in a 50/50 split. 

We train for up to 50 epochs, using Adam optimizer with learning rate set to 0.001. When we use pretrained word embeddings (either debiased or not), we do not finetune or update them. We do not use any L2 regularization or dropout, instead employing early stopping with patience of 5 epochs, using validation loss as the stopping criterion. We use batch size 256. Training is completed on 1 NVIDIA Titan Xp with 12 GB of memory.  

\subsection{Ensemble Model}
We use scikit-learn \cite{scikit-learn} with the default setting of L2 regularization with $C=1$

\section{Sets of social-specific Words}
\label{appendix:c}

\subsection{Sets of Gender-specific Words}
\begin{itemize}
\setlength{\itemsep}{-0.1pt}
  \item \{"he", "she"\}
  \item	\{"his", "hers"\}
  \item	\{"son", "daughter"\}
  \item	\{"father", "mother"\}
  \item	\{"male", "female"\}
  \item	\{"boy", "girl"\}
  \item	\{"uncle", "aunt"\}
\end{itemize}
\subsection{Sets of Racial-specific Words}
\begin{itemize}
\setlength{\itemsep}{-0.1pt}
    \item \{"black", "caucasian", "asian", "hispanics"\}
	\item \{"african", "caucasian", "asian", "hispanics"\}
	\item \{"black", "white", "asian", "hispanics"\}
	\item \{"africa", "america", "asia", "hispanics"\}
	\item \{"africa", "america", "china", "hispanics"\}
	\item \{"africa", "europe", "asia", "hispanics"\}
	\item \{"black", "caucasian", "asian", "latino"\}
	\item \{"african", "caucasian", "asian", "latino"\}
	\item \{"black", "white", "asian", "latino"\}
	\item \{"africa", "america", "asia", "latino"\}
	\item \{"africa", "america", "china", "latino"\}
	\item \{"africa", "europe", "asia", "latino"\}
	\item \{"black", "caucasian", "asian", "spanish"\}
	\item \{"african", "caucasian", "asian", "spanish"\}
	\item \{"black", "white", "asian", "spanish"\}
	\item \{"africa", "america", "asia", "spanish"\}
	\item \{"africa", "america", "china", "spanish"\}
	\item \{"africa", "europe", "asia", "spanish"\}
\end{itemize}

\section{Hard Debiasing}
\label{appendix:d}
Hard debiasing is a debiasing algorithm which involves two steps: neutralize and equalize. Neutralization ensures that all the social-neural words in the social subspace do not contain bias (e.g. doctors and nurses). Equalization forces that social-specific words are equidistant to all words in each equality set (e.g. the bias components in man and woman are in opposite directions but with same magnitude) \cite{bolukbasi2016man, manzini-etal-2019-black}. Following \citet{manzini-etal-2019-black}, hard debiasing is formulated as follows: given a bias social subspace $\mathcal{B}$ spanned by the vectors $\{b_{1}, b_{2}, ..., b_{n}\}$, the embedding of a word in this subspace is:
\[w_\mathcal{B} = \sum_{i=1}^{n} \langle{w}, b_{i}\rangle{b_{i}}\]
To neutralize, each word $w \in{N}$, where $N$ is the set of social-neural words, remove the bias components from the word and the re-embedded word $\overrightarrow{w}$ is obtained as: 
\[\overrightarrow{w} = \frac{w - w_\mathcal{B}}{\parallel{w - w_\mathcal{B}}\parallel}\]
To equalize, for an equality set $E$, let $\mu$ be the mean embeddings of the equlity set $E$, which is defined as:
\[\mu = \frac{w}{E}\sum_{w\in{E}}\]
For each word $w \in{E}$, the equalization is defined as:
\[\hat{w} = (\mu - \mu_{\mathcal{B}}) + \sqrt{1 - \parallel{\mu - \mu_{\mathcal{B}}\parallel^2}} \frac{w - w_\mathcal{B}}{{\parallel{w - w_\mathcal{B}}}\parallel}\]
When doing racial debiasing, we divide ethnicity into groups: White, Black, Asian, and Hispanics. We do not contain the "other" group as it hard to define social-specific sets and analogies for "other".  

\section{Phenotyping Task}
\label{sec:appendix:phenotyping}

In Figure \ref{fig:big_plot} we plot performance and fairness for the phenotyping task, specifically the detection of disorders of lipid metabolism. This task was selected as it is the phenotyping task with the most balanced labels with 16855 negative instances and 12239 positive instances in the training data. Thus, it should be more amenable to EO postprocessing. As expected we see that EO postprocessing succeeds in reducing the TP/TN ranges at the cost of AUC. We also again see that ensembling improves performance both before and after postprocessing. For this task specifically we observe that using debiased word embeddings improves AUC compared to the non-debiased word embeddings.

\section{Full Results}
Our experiment universe consisted of the cross product between choice of protected attribute (gender, ethnicity, insurance status), task (phenotyping, in-hospital mortality prediction, decompensation), hard vs soft EO postprocessing and word embedding vs debiased word embedding. 

\label{sec:appendix:full_result}
\subsection{Fairness/Performance on the In-Hospital Mortality Task}
\begin{figure*}[!ht]
  \centering
  \includegraphics[width=\linewidth]{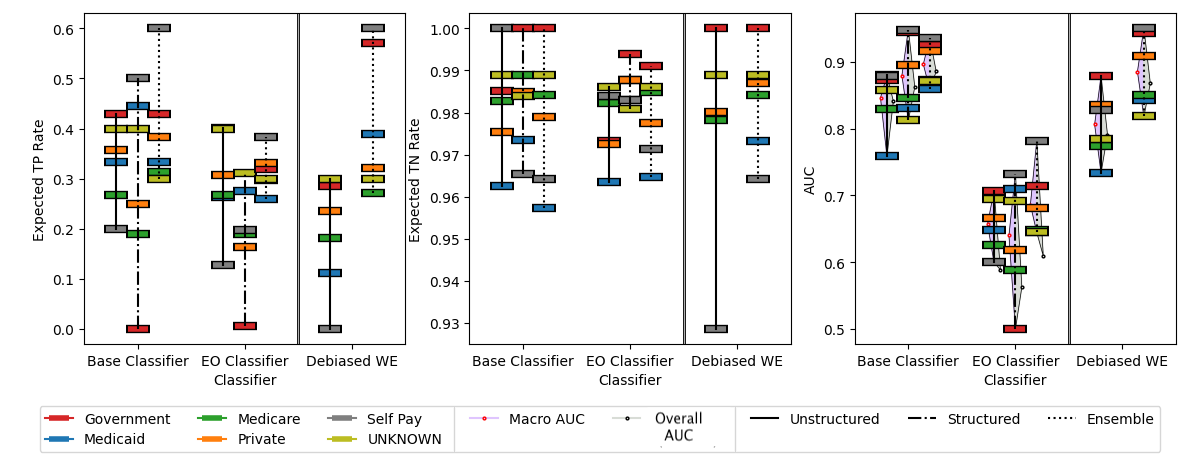}
  \caption{Plot of Fairness and Performance on the in-hospital mortality task. Note that debiased word embeddings slightly worsen the TP gap in this task (left most graph), while improving the TN gap (middle graph). EO reduces both gaps, at a major cost in performance (right most graph).  }\label{fig:ihm_big_plot}
\end{figure*}

We provide a more detailed set of graphs for an in-hospital mortality prediction task, where we used hard EO postprocessing on protected groups defined by insurance status. We illustrate the TP/TN/AUC metrics for each protected group in Figure \ref{fig:ihm_big_plot}. 

In this task configuration, as well as the task configuration in Figure \ref{fig:big_plot} EO postprocessing is applied to hard classification of the three classifiers in the Base Classifier column, to produce the EO Classifier column. The Debiased Word Embedding (WE) column contains an unstructured classifier using word embeddings debiased for 4 ethnicities, and an ensemble created by merging the aforementioned classifier with the structured base classifier. We utilize debiasing on ethnicity type as a proxy for insurance status, as mentioned in the Discussion.

Note that EO post-processing sometimes \textit{worsens} the TP/TN spread, as in the TP graph for the structured classifier. We therefore qualify our EO results by noting the limitations of our real-world dataset, which include significant group and label imbalance and non-binary group labels, all of which impact the results of EO post-processing (see Appendix \ref{sec:appendix:cohort_stats}). 

Finally, on this task configuration, we observe that debiased word embeddings are not a panacea.  We note that WED has slightly worsened the TP gap, and does not offer a clear cut performance improvement as on the phenotyping task M. Therefore, further research is needed to explore when and why debiased word embeddings may simultaneously improve fairness and performance. Ultimately, domain expertise and focus on the downstream impact on the patient experience will be critical for leveraging any of these fair machine learning models in clinical applications.

\subsection{Full table of results}
The performance for all model and tasks tried can be found in Table \ref{tab:full_perf}. Note that debiased word embeddings can improve the performance (micro and macro AUC), even compared to an unconstrained classifier using clinically relevant BioWordVecembeddings.

\begin{table}[htb]
\centering
\resizebox{\columnwidth}{!}{
\begin{tabular}{clllllll}
\cline{2-7}
\multicolumn{1}{l|}{} &
  \multicolumn{2}{c|}{IHM} &
  \multicolumn{2}{c|}{Phenotyping} &
  \multicolumn{2}{c|}{Decompen.} &
  \multicolumn{1}{l}{} \\ \cline{1-7}
\multicolumn{1}{|l|}{} &
  \multicolumn{1}{c|}{\begin{tabular}[c]{@{}c@{}}AUC \\ PRC\end{tabular}} &
  \multicolumn{1}{c|}{\begin{tabular}[c]{@{}c@{}}AUC \\ ROC\end{tabular}} &
  \multicolumn{1}{c|}{\begin{tabular}[c]{@{}c@{}}Macro \\ AUC \\ ROC\end{tabular}} &
  \multicolumn{1}{c|}{\begin{tabular}[c]{@{}c@{}}Micro \\ AUC \\ ROC\end{tabular}} &
    \multicolumn{1}{c|}{\begin{tabular}[c]{@{}c@{}}AUC \\ PRC\end{tabular}} &
  \multicolumn{1}{c|}{\begin{tabular}[c]{@{}c@{}}AUC \\ ROC\end{tabular}} &
   \\ \cline{1-7}
   \multicolumn{1}{|c|}{\begin{tabular}[c]{@{}c@{}}Harutyunyan \\ et. al\\  (2019) \\ -- No Text\end{tabular}}& \multicolumn{1}{l|}{0.515} & \multicolumn{1}{l|}{0.862} &  
\multicolumn{1}{c|}{0.776} &
\multicolumn{1}{c|}{0.825} &  
\multicolumn{1}{c|}{0.344} &
\multicolumn{1}{c|}{0.911} &  \\
\cline{1-7}
\multicolumn{1}{|c|}{\begin{tabular}[c]{@{}c@{}}Khadanga \\ et. al \\  (2019) \\ -- Ensemble\end{tabular}}    & \multicolumn{1}{l|}{0.525} & \multicolumn{1}{l|}{0.865} &
\multicolumn{1}{c|}{--} & 
\multicolumn{1}{c|}{--} &
\multicolumn{1}{c|}{0.345} & 
\multicolumn{1}{c|}{0.907} &\\ \cline{1-7}
\multicolumn{1}{|c|}{\begin{tabular}[c]{@{}c@{}}Ours \\ -- Text Only\end{tabular}} &
  \multicolumn{1}{l|}{0.472} &
  \multicolumn{1}{l|}{0.815} &
\multicolumn{1}{c|}{0.766} &  
\multicolumn{1}{c|}{0.829} & 
\multicolumn{1}{c|}{0.235} &  
\multicolumn{1}{c|}{0.867} &\\ \cline{1-7}
   \multicolumn{1}{|c|}{\begin{tabular}[c]{@{}c@{}}Ours \\ -- Text Only \\ + BioWordVec \end{tabular}} &
  \multicolumn{1}{l|}{0.489} &
  \multicolumn{1}{l|}{0.841} &
\multicolumn{1}{c|}{0.771} & 
\multicolumn{1}{c|}{0.837} &  
\multicolumn{1}{c|}{0.225} & 
\multicolumn{1}{c|}{0.879} &\\ \cline{1-7}
   \multicolumn{1}{|c|}{\begin{tabular}[c]{@{}c@{}}Ours \\ -- Text Only \\ + BioWordVec \\ + Debiasing \end{tabular}} &
  \multicolumn{1}{l|}{0.392} &
  \multicolumn{1}{l|}{0.790} &
\multicolumn{1}{c|}{0.831} & 
\multicolumn{1}{c|}{0.874} &  
\multicolumn{1}{c|}{0.265} & 
\multicolumn{1}{c|}{0.331} & \\ \cline{1-7}
\multicolumn{1}{|c|}{\begin{tabular}[c]{@{}c@{}}Ours \\ -- Ensemble\end{tabular}}           & \multicolumn{1}{l|}{\textbf{0.582}} & 
\multicolumn{1}{l|}{0.880} &  
\multicolumn{1}{c|}{0.822} &
\multicolumn{1}{c|}{0.861} & 
\multicolumn{1}{c|}{0.399} &
\multicolumn{1}{c|}{0.917} & \\ \cline{1-7}
\multicolumn{1}{|c|}{\begin{tabular}[c]{@{}c@{}}Ours \\ -- Ensemble \\ + BioWordVec\end{tabular}}           & \multicolumn{1}{l|}{\textbf{0.582}} & \multicolumn{1}{c|}{\textbf{0.886}} &  
\multicolumn{1}{c|}{0.829} & 
\multicolumn{1}{c|}{0.870} & 
\multicolumn{1}{c|}{\textbf{0.404}} & 
\multicolumn{1}{c|}{\textbf{0.920}} & \\ \cline{1-7}
\multicolumn{1}{|c|}{\begin{tabular}[c]{@{}c@{}}Ours \\ -- Ensemble \\ + BioWordVec \\ + Debiasing\end{tabular}}           & 
\multicolumn{1}{l|}{0.539} & 
\multicolumn{1}{c|}{0.870} &  
\multicolumn{1}{c|}{\textbf{0.854}} & 
\multicolumn{1}{c|}{\textbf{0.888}} & 
\multicolumn{1}{c|}{0.405} & 
\multicolumn{1}{c|}{0.922} & \\ \cline{1-7}
\end{tabular}}
\caption{Leveraging clinical pretrained word embeddings improves performance compared to training word embeddings from scratch in the text-only model. Ensembling the text-only model with the clinical time series classifier improves performance further. As with \citet{khadanga2019using}, our results are not directly comparable with \citet{Harutyunyan2019} since we ignore patients without any clinical notes.
 }
\label{tab:full_perf}
\end{table}

\end{document}